\documentclass[lettersize,journal]{template/IEEEtran}
\IEEEoverridecommandlockouts

\usepackage[numbers,sort&compress]{natbib}
\usepackage{booktabs}
\usepackage{amsmath}
\usepackage{amssymb}
\usepackage{graphicx}
\usepackage{xcolor}
\usepackage{multirow}
\usepackage{algorithm}
\usepackage{algpseudocode}
\usepackage{tikz}
\usepackage{url}
\newcommand{\nDiag}{688}
\newcommand{\nRep}{600}
\newcommand{\collClean}{16.1}
\newcommand{\collGlThreeOn}{19.9}

\newcommand{\pGlThreeOn}{$<$0.001}

\newcommand{\dcollGlThreeOff}{-0.4}

\newcommand{\collGlFiveOn}{20.9}

\newcommand{\collGlFiveSudden}{16.6}

\newcommand{\pGlFiveSudden}{0.784}

\newcommand{\collOcOn}{20.9}

\newcommand{\pOcOn}{$<$0.001}

\newcommand{\dcollOcOff}{+0.7}

\newcommand{\dcollTunnel}{+0.4}

\newcommand{\collFog}{20.2}

\newcommand{\pFog}{$<$0.001}

\newcommand{\dcollRain}{+0.4}

\newcommand{\dcollRerun}{-0.1}

\newcommand{\fastOcOn}{92.3}
\newcommand{\dprogNewOcOn}{5.4}
\newcommand{\fastGlFiveOn}{86.4}

\newcommand{\repOcOnNone}{21.0}
\newcommand{\repOcOnTrf}{18.2}

\newcommand{\repOcOnFreeze}{15.5}

\newcommand{\repOcOnClahe}{19.5}

\newcommand{\repOcOnDcp}{19.8}

\newcommand{\repOcOnInpaint}{20.7}

\newcommand{\repOcOnNoAlign}{17.5}

\newcommand{\repOcOnSat}{20.3}

\newcommand{\repOcOnTex}{18.5}

\newcommand{\repOcOnAdapt}{16.0}
\newcommand{\repPOcOnAdapt}{$<$0.001}

\newcommand{\repGlFiveOnNone}{20.8}
\newcommand{\repGlFiveOnTrf}{19.0}

\newcommand{\repGlFiveOnNoAlign}{17.0}

\newcommand{\repGlFiveOnSat}{17.2}

\newcommand{\repGlFiveOnTex}{19.1}

\newcommand{\repGlFiveOnAdapt}{16.0}
\newcommand{\repPGlFiveOnAdapt}{$<$0.001}

\newcommand{\repFogDcp}{15.8}

\newcommand{\repFogTex}{16.5}

\newcommand{\repFogAdapt}{17.2}

\newcommand{\cleanNone}{16.3}
\newcommand{\cleanAdapt}{14.8}
\newcommand{\cleanNewAdapt}{5}
\newcommand{\cleanResAdapt}{14}

\newcommand{\cleanDetFreeze}{17.0}
\newcommand{\cleanNewDetFreeze}{19}
\newcommand{\cleanResDetFreeze}{15}

\newcommand{\cleanFarther}{99.1}
\newcommand{\cleanFirstMed}{4.5}
\newcommand{\cleanAfterThree}{96.3}
\newcommand{\gtOriented}{0.0}

\newcommand{\onoffAvOc}{+0.66}
\newcommand{\onoffPAvOc}{$<$0.001}

\newcommand{\onoffQdOc}{-1.12}
\newcommand{\onoffPQdOc}{$<$0.001}
\newcommand{\onoffQdGl}{-0.69}
\newcommand{\onoffPQdGl}{$<$0.001}

\newcommand{\repQdOcBefore}{3.3}
\newcommand{\repQdOcAfter}{4.2}

\newcommand{\repQdGlBefore}{3.2}
\newcommand{\repQdGlAfter}{4.3}

\newcommand{\repQdFogBefore}{11.7}
\newcommand{\repQdFogAfter}{4.0}
\newcommand{\repQdFogP}{$<$0.001}
\newcommand{\repQdClBefore}{2.7}
\newcommand{\repQdClAfter}{3.2}
\newcommand{\repQdClP}{0.453}
\newcommand{\diagQdFog}{+8.7}
\newcommand{\diagPQdFog}{$<$0.001}
\newcommand{\diagQdOc}{+0.6}
\newcommand{\diagPQdOc}{0.541}
\newcommand{\onoffImOc}{+0.17}
\newcommand{\onoffPImOc}{$<$0.001}

\newcommand{\nIm}{134}

\newcommand{\wNAvRain}{186}

\newcommand{\wTrfBeforeAvNight}{22.9}
\newcommand{\wTrfAfterAvNight}{19.2}
\newcommand{\wTrfPAvNight}{$<$0.001}
\newcommand{\wOnOffAvNight}{+4.4}
\newcommand{\wOnOffPAvNight}{$<$0.001}
\newcommand{\wOnOffProgAvNight}{+2.75}
\newcommand{\wNAvNight}{1040}

\newcommand{\wTrfBeforeQdNight}{13.9}
\newcommand{\wTrfAfterQdNight}{7.0}
\newcommand{\wTrfPQdNight}{$<$0.001}
\newcommand{\wOnOffQdNight}{+9.0}
\newcommand{\wOnOffPQdNight}{$<$0.001}
\newcommand{\wOnOffProgQdNight}{-0.51}

\newcommand{\hoAvNCl}{595}

\newcommand{\hoAvOcNone}{25.6}

\newcommand{\hoAvOcPro}{24.2}

\newcommand{\hoAvOcAdapt}{22.7}

\newcommand{\hoAvOcG}{21.5}
\newcommand{\hoAvOcGP}{$<$0.001}
\newcommand{\hoAvNGl}{585}

\newcommand{\hoAvGlMoce}{23.2}

\newcommand{\hoAvGlG}{23.4}

\newcommand{\hoQdClFreeze}{6.2}

\newcommand{\hoQdClFreezePG}{$<$0.001}

\newcommand{\hoQdClG}{4.3}

\newcommand{\hoQdNOc}{599}
\newcommand{\hoQdOcNone}{13.4}

\newcommand{\hoQdOcAdair}{11.5}

\newcommand{\hoQdOcFlare}{13.2}

\newcommand{\hoQdOcG}{7.3}

\newcommand{\hoQdGlNone}{11.5}

\newcommand{\hoQdGlG}{7.7}
\newcommand{\hoQdGlGP}{0.001}

\newcommand{\hoQdFogFreeze}{7.2}

\newcommand{\hoQdFogG}{4.2}

\newcommand{\hoAvgNone}{15.8}

\newcommand{\hoAvgDcp}{14.9}

\newcommand{\hoMsAdair}{214}
\newcommand{\hoAvgMoce}{15.3}
\newcommand{\hoMsMoce}{194}

\newcommand{\hoAvgG}{13.7}
\newcommand{\hoMsG}{127}

\def\tArea{37}
\def\mArea{61}

\newcommand{\bench}{HIDE}
\newcommand{\method}{TRF}
\newcommand{\model}{AutoVLA}

\begin{document}
\bstctlcite{IEEEexample:BSTcontrol}

\title{Hidden Hazards: Placement-Controlled Camera Degradation and
Training-Free Temporal Repair for Driving VLA Models}

\author{Zikun~Guo and Chongchong~Song%
\thanks{(Corresponding author: Zikun Guo.)}%
\thanks{Zikun Guo is with the School of Autonomous Driving, Xinwei Institute
of Artificial Intelligence, China (e-mail: gzk798412226@gmail.com).}%
\thanks{Chongchong Song is with Geely Automobile Research Institute (Ningbo)
Co., Ltd., Ningbo, China (e-mail: songchongchong@geely.com).}}

\markboth{Preprint}{Guo and Song: Placement-Controlled Camera Degradation and Training-Free Temporal Repair}

\maketitle

\begin{abstract}
Camera degradation in driving often affects only part of the view. Low sun,
oncoming headlights and dirt on the lens degrade one region of one camera,
while the rest of the image remains clear. Existing robustness benchmarks
apply each corruption to the whole frame and therefore conflate image
degradation with the occlusion of a road user. This paper introduces
\bench{}, a hazard indexed degradation benchmark for end-to-end driving.
\bench{} projects each annotated hazard on the driven path into the camera
image. Glare or lens occlusion is then placed either over the hazard or at a
matched position away from the hazard. Tunnel exit overexposure, fog and rain
complete the benchmark, each with persistent and sudden onset. Experiments
with three driving vision-language-action models on nuScenes show that
concealing the hazard alters the plan, whereas the same degradation at a
matched position leaves the plan unchanged. AutoVLA and Impromptu-VLA maintain or increase
speed when the hazard is concealed, and AutoVLA shows a higher collision rate.
Qwen-Drive reduces speed under local degradation but shows a higher collision
rate under fog. Degradation that
preserves clean frames in the input history has little effect on safety, and
the main effects persist in real night recordings. Based on these results,
this paper proposes \method{}, a training-free temporal reference fill.
\method{} detects degraded regions relative to the last clean frame, refills
small regions from this reference and substitutes the reference for frames
that are largely lost. \method{} reduces the collision rate under the
conditions that degrade each planner, with no significant increase on clean
input in daylight, rain and night. On held-out scenes, a \method{}
variant with a dehazing branch achieves a lower collision rate under hazard
occlusion than recent learned restoration, flare removal and video inpainting
models. A clean temporal reference is therefore an effective training-free
repair for camera-based driving planners.

\end{abstract}


\section{Introduction}

\IEEEPARstart{E}{nd-to-end} driving models map camera frames and the ego state to a future
trajectory \citep{uniad,vad}. Vision-language-action (VLA) models extend this
design with a pretrained vision-language backbone and a language instruction
\citep{autovla,opendrivevla,impromptuvla}. In the camera-based planners of both
families, the cameras are the only source of information about other road
users. A degraded camera image therefore leaves the planner without a second
source of information about the scene.

In real driving, camera degradation is rarely uniform. A low sun or an oncoming
high beam produces a bright bloom around one point of the image. Leaving a
tunnel overexposes the frame for a fraction of a second before the auto
exposure adapts. Mud or a water film covers part of one lens. In each case one
region of one camera is lost, while the rest of the view remains clear
\citep{adverseweather,glaredataset}.

Current robustness benchmarks omit this spatial structure. These benchmarks
apply a corruption to the whole frame at a set severity
\citep{imagenetc,robobev,robodrivevlm}, or perturb the sensor stream and the
ego state at the system level \citep{bench2driverobust}. Such protocols conflate
two causes of failure: an overall loss of image quality, and the concealment of
the road user that constrains the plan. Only the second cause depends on where
the degradation falls.

We therefore treat the placement of the degradation as a controlled variable.
We introduce \bench{}, a hazard-indexed degradation evaluation. For every scene
with a road user on the driven path, \bench{} projects the hazard into the
camera image. Glare or lens occlusion is then placed either over the hazard or
at a matched position away from the hazard. The two placements share the
camera, the image row and the severity. The placement difference
isolates hazard concealment from the overall loss of image quality. Tunnel exit,
fog and rain, with persistent and sudden onset, complete the protocol. Safety
is scored against the annotated future track of every hazard.

We apply \bench{} to three released driving VLAs on nuScenes
\citep{nuscenes}: AutoVLA \citep{autovla}, Qwen-Drive \citep{qwendrive} and
Impromptu-VLA \citep{impromptuvla}. Four results follow. First, concealing the
hazard alters the plan, whereas the same degradation at a matched position away
from the hazard leaves the plan unchanged. Second, the planners respond in
opposite directions. AutoVLA and Impromptu-VLA maintain or increase speed when
the hazard is concealed, and the collision rate of AutoVLA rises. Qwen-Drive
reduces speed. Third, Qwen-Drive shows a higher collision rate under fog, which
lowers the contrast of the whole view without removing any single region.
Fourth, a sudden degradation that preserves clean frames in the input history
has little effect on safety. On real rain and night recordings, concealing the
hazard has the same effect. At night, concealment raises the collision rate of
both planners tested on these recordings.

\begin{figure*}[t]
  \centering
  \includegraphics{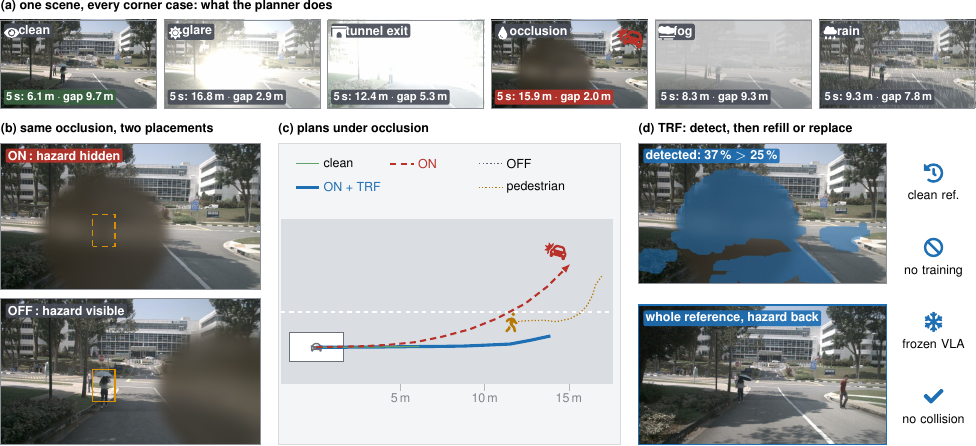}
  \caption{One scene from \bench{}, with a pedestrian 11.6\,m ahead.
  \textbf{(a)} The scene under each degradation family, with the 5\,s
  progress of the plan and the minimum distance between the plan and the
  pedestrian. Every
  degradation increases planned progress relative to the clean plan.
  \textbf{(b)} The same lens occlusion covers the pedestrian (on) or the far
  side of the frame (off). \textbf{(c)} Plans of the driving VLA. Under the on
  placement, the plan accelerates and reaches the pedestrian. Under the off
  placement, the plan remains close to the clean plan. \textbf{(d)} \method{} flags \tArea{}\% of the frame as degraded. Above a
  quarter of the image, \method{} passes the clean reference frame. The
  pedestrian becomes visible again, and the plan stops short of the
  pedestrian.}
  \label{fig:overview}
\end{figure*}

The fourth result motivates a repair that requires no training. We propose
\method{}, a temporal reference fill. \method{} stores the last frame known to
be clean as a reference. \method{} detects degraded regions by saturation and
loss of texture relative to this reference. A small region is refilled from
the reference, and the whole reference replaces the frame when more than a
quarter of the frame is lost. The planner remains unchanged.
Fig.~\ref{fig:overview} summarises the benchmark, the failure mode that the benchmark exposes,
and the repair.

Our contributions are as follows.
\begin{itemize}
  \item \textbf{A placement-controlled benchmark.} \bench{} indexes camera
  degradation to the in-path hazard with matched on and off placements. \bench{}
  covers 691 nuScenes scenes, five degradation families and two onset patterns,
  and checks collisions against annotated future tracks.
  \item \textbf{Two failure modes of driving VLAs.} On three VLAs, concealing the hazard leads two planners to
  maintain or increase speed and Qwen-Drive to reduce speed. Qwen-Drive
  shows a higher collision rate under fog. Matched off-hazard degradation and
  sudden onset with clean history have little effect on safety.
  \item \textbf{A training-free repair.} \method{} refills a small degraded
  region or replaces a largely lost frame with the last clean frame. \method{}
  lowers the collision rate under the conditions that degrade each of the three
  planners, including on real night recordings. The planners remain unchanged,
  and the collision rate on clean input shows no significant increase. On a held-out split, \method{} is compared with
  recent learned restoration and video inpainting models under a protocol fixed
  in advance.
\end{itemize}

\section{Related Work}

\paragraph{End-to-end driving and VLAs}
Modular end-to-end planners such as UniAD \citep{uniad} and VAD \citep{vad}
learn perception, prediction and planning jointly from multi-view cameras.
Driving VLA models replace the task-specific stack with a pretrained
vision-language backbone that processes camera frames, the ego state and an
instruction, and decodes a trajectory \citep{autovla,opendrivevla,impromptuvla,senna}.
Language models also read street-level imagery well enough to rate urban
scenes \citep{han2026street}, which supports using a vision-language backbone
for driving. Open-loop displacement error on nuScenes is dominated by the ego
state \citep{egostatus}, so we score safety against the annotated future of each
hazard, paired with the same scene without degradation.

\paragraph{Reliability of multimodal models}
Multimodal models can follow cues that are not evidence. Video language models
change correct answers under user pressure \citep{zhou2026flattery}. Medical
vision-language models defer to social cues over the image, and filtering those
cues restores accuracy \citep{xu2025medsyco}. Driving planners rely on agents
that co-occur with the expert action without causing the action \citep{cadet}. These
studies perturb what the model is told or what the scene contains. \bench{}
conceals visual information and measures whether the plan accounts for the
loss.

\paragraph{Corruption robustness benchmarks}
ImageNet-C introduced severity graded common corruptions for image
classification \citep{imagenetc}. RoboBEV applies the same idea to
bird's-eye-view perception, with corruptions from sensor failure, motion,
lighting and weather \citep{robobev}. RoboDriveBench extends corruption testing
to vision-language driving models that output trajectories \citep{robodrivevlm}.
Bench2Drive-Robust evaluates closed-loop driving under camera stream failures,
ego state errors and control delay \citep{bench2driverobust}. Fog is commonly
rendered with an atmospheric scattering model over a depth estimate
\citep{foggycityscapes}. Benchmarks of this kind are synthesised under control, which is also how
generative models produce labelled data where real data is scarce
\citep{guo2025cag}. Vision transformers trained for field imagery face the
same shift in illumination \citep{guo2025crop}. All of these protocols apply a
corruption to the whole frame, or to whole cameras, and report the drop in
performance. \bench{} retains
the corruption families and controls where in the image the corruption falls
relative to the hazard.

\paragraph{Robustness through sensor fusion}
A second line of work makes the model itself robust to a failed or degraded
sensor. MetaBEV and UniBEV maintain camera-LiDAR detection when one modality
is missing \citep{metabev,unibev}. SAMFusion weights modalities by distance and
weather \citep{samfusion}. VG-SAF gates camera-LiDAR fusion with dense
reliability estimates and evaluates end-to-end driving under asymmetric sensor
degradation \citep{vgsaf}. These methods require retraining with a second
modality. \method{} uses the history of the same camera at test time.

\begin{figure*}[t]
  \centering
  \includegraphics{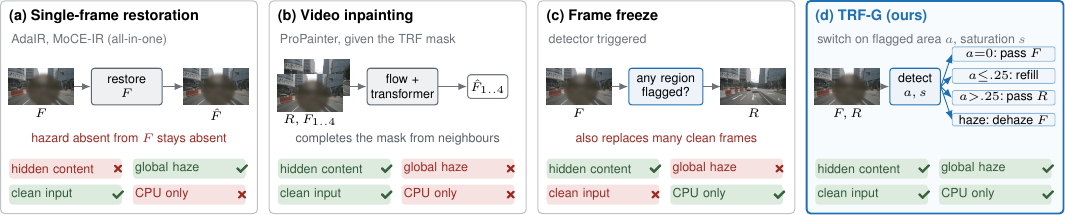}
  \caption{Repair paradigms placed in front of an unchanged planner. Each chip
  states whether the paradigm restores content that the current frame has lost,
  handles a global loss of contrast, leaves clean input safe, and runs without a
  GPU. Every chip follows from Table~\ref{tab:heldout}.}
  \label{fig:paradigms}
\end{figure*}

\paragraph{Input restoration and lightweight models}
Classical restoration removes haze with the dark channel prior \citep{dcp},
equalises contrast locally \citep{clahe}, or inpaints missing regions from
their surroundings \citep{telea}. Recent all-in-one networks restore several
degradations with one model \citep{adair,moceir}, flare removal networks target
lens flare at night \citep{flare7kpp}, and video inpainting completes masked
regions from neighbouring frames \citep{propainter}.
Fig.~\ref{fig:paradigms} contrasts these repair paradigms with \method{}-G.

These restoration methods operate independently of the
downstream planning objective. Restoring a degraded signal before a fixed
downstream model is also the design of real-time speech enhancement for
hearing aids \citep{fan2025tsdcaba}. A repair that runs in front of a planner
on the vehicle must have low computational cost. Compressed question answering models
\citep{guo2023idsextract} and cluster-aggregated lightweight transformers
\citep{guo2025cat} illustrate model size reduction. \method{} adds no
learned parameters at all.

\paragraph{Planning under constraints}
Collision-free planning is studied well beyond driving. Multi-robot path
planning balances exploration with structural priors \citep{guo2026structural}.
Task allocation and scheduling for robot teams use clustering
\citep{adedigba2025ivec} and cooperative coevolution \citep{adedigba2025ccga},
and transformer policies trained with reinforcement learning solve routing
problems \citep{guo2025dyt}. These planners assume complete perception of obstacles.

\paragraph{Test-time repair of planners}
CADET audits and repairs spurious reliance in pretrained planners at inference
time, using a physics prior as an external reference \citep{cadet}. \method{}
follows the same training-free principle, with the last clean frame as the
reference and the input image as the target.

\section{The \bench{} Benchmark}
\label{sec:benchmark}

\paragraph{Reference planner}
The benchmark is defined on the input of \model{} \citep{autovla}, a driving VLA built on Qwen2.5-VL-3B
\citep{qwen25vl}. At each keyframe the model reads the front, front-left and
front-right cameras, four frames each at 2\,Hz (the current frame and the
preceding 1.5\,s). The model also receives the ego velocity, the ego acceleration and a
navigation command (\texttt{KEEP FORWARD}, \texttt{TURN LEFT} or
\texttt{TURN RIGHT}). The model outputs ten poses at 2\,Hz, covering 5\,s. We give the
model the command that matches the manoeuvre the human driver performed, so the
clean run is the standard open-loop setting. The same scenes and frames are applied to two further planners in Section~\ref{sec:planners}.

\begin{figure*}[t]
  \centering
  \includegraphics{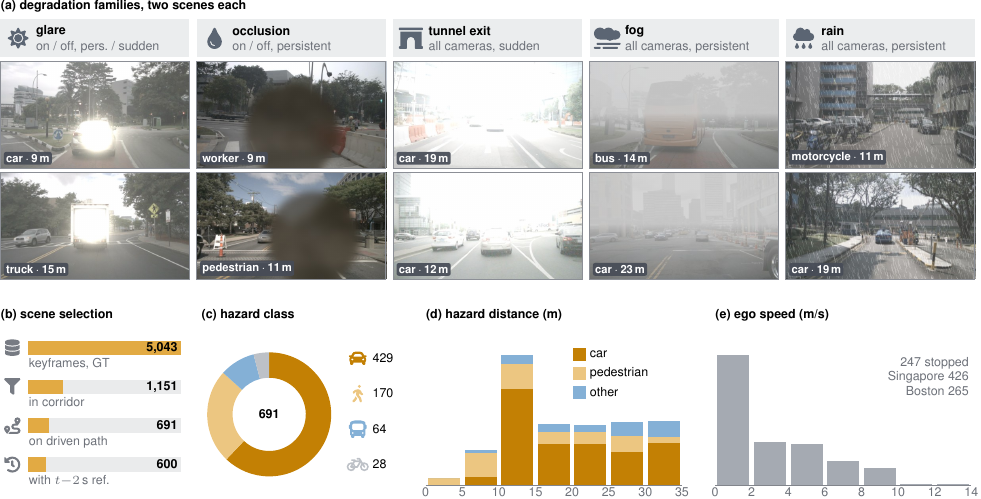}
  \caption{The \bench{} dataset. \textbf{(a)} The five degradation families,
  each shown on two scenes, with the nearest hazard and its distance.
  \textbf{(b)} Scene selection from nuScenes keyframes with a 5\,s
  ground-truth future. \textbf{(c)} Class of the nearest hazard.
  \textbf{(d)} Hazard distance by class. \textbf{(e)} Ego speed at the
  keyframe.}
  \label{fig:dataset}
\end{figure*}

\paragraph{In-path hazard scenes}
Scenes come from nuScenes \citep{nuscenes}. A keyframe is kept when a vehicle,
pedestrian or cyclist lies ahead of the ego vehicle within 35\,m and within
2\,m of the ego heading line. Roads curve, so this corridor alone admits parked agents that the vehicle never approaches. We therefore also require the
nearest hazard to lie within 2.5\,m of the path the human driver actually took
over the next 5\,s. The hazard must also project into the front camera. These criteria
yield 691 scenes from 80 drives in Singapore and Boston
(Fig.~\ref{fig:dataset}b). Cars are the nearest hazard in 429 scenes and
pedestrians in 170. The median distance to the nearest hazard is 18.8\,m
(Fig.~\ref{fig:dataset}c,d).

\paragraph{Degradation families}
Each family is a function of the image and a severity $s \in \{1,\dots,5\}$.
\begin{itemize}
  \item \textbf{Glare} models a point light source. A narrow Gaussian core and
  a heavy-tailed bloom surround the source. A veiling term lifts the black level
  of the whole frame, and faint ghosts lie on the line through the image centre.
  Values above the sensor range are clipped.
  \item \textbf{Tunnel exit} models a frame exposed for a dark interior. A
  soft-knee tone curve with a gain of $1+4s/5$ overexposes the frame, with the
  strongest lift at the opening in the upper centre.
  \item \textbf{Occlusion} models dirt or a water film on the lens. A blurred,
  darkened disc covers a fraction $0.08+0.13(s-1)$ of the image.
  \item \textbf{Fog} follows the scattering model
  $I = J\,t + A\,(1-t)$ with $t=\exp(-\beta d)$ \citep{foggycityscapes}. The
  depth $d$ is a ground-plane proxy from the image row, capped above the
  horizon, so the protocol needs no depth sensor.
  \item \textbf{Rain} adds motion-blurred streaks and a mild veil.
\end{itemize}

\paragraph{Hazard-indexed placement}
We project the centre of the nearest hazard into each camera with the nuScenes
calibration. Glare and occlusion are then placed in one of two ways. The on placement centres the degradation on the projected hazard. The off placement preserves the camera, the image row and the severity, and
moves the degradation to the far side of the frame (horizontal position 0.15 or
0.85). The pair differs only in whether the hazard remains visible. The on minus
off difference therefore isolates the effect of losing the hazard from the
effect of degrading the image. Tunnel exit, fog and rain are global and are
applied to all three cameras. Every family at three severities, together with the matched off placement, is shown in Fig.~\ref{fig:gallery} of the supplementary material.

\paragraph{Onset}
A persistent degradation is applied to all four frames of the input
window. A sudden degradation is applied to the current frame only,
which models the moment of leaving a tunnel or turning into the sun. The three
earlier frames then remain clean.

\paragraph{Metrics}
Every metric is computed per scene and compared with the same scene without
degradation.
\begin{itemize}
  \item \textbf{Collision rate.} The ego box ($4.08\times1.73$\,m) at each
  planned pose is tested for overlap with every hazard box at that agent's
  annotated position at the same time. Hazard motion therefore follows the recorded
  tracks. The human trajectory has a collision rate of 0\% under this check.
  We report the number of scenes that collide only under the degradation
  (new) and only without degradation (resolved), with an exact McNemar test.
  \item \textbf{Progress} is the distance covered by the plan over its horizon
  (5\,s for \model{}). A
  negative change indicates reduced planned progress.
  \item \textbf{Clearance} is the minimum distance between the planned poses
  and the hazard track.
  \item \textbf{L2} is the mean displacement error to the human trajectory at
  1, 2 and 3\,s.
\end{itemize}
\model{} decodes with sampling. Every clean scene is therefore run twice. The
difference between two clean runs gives the noise floor of each metric.

\section{\method{}: Temporal Reference Fill}
\label{sec:method}

\begin{figure*}[t]
  \centering
  \includegraphics{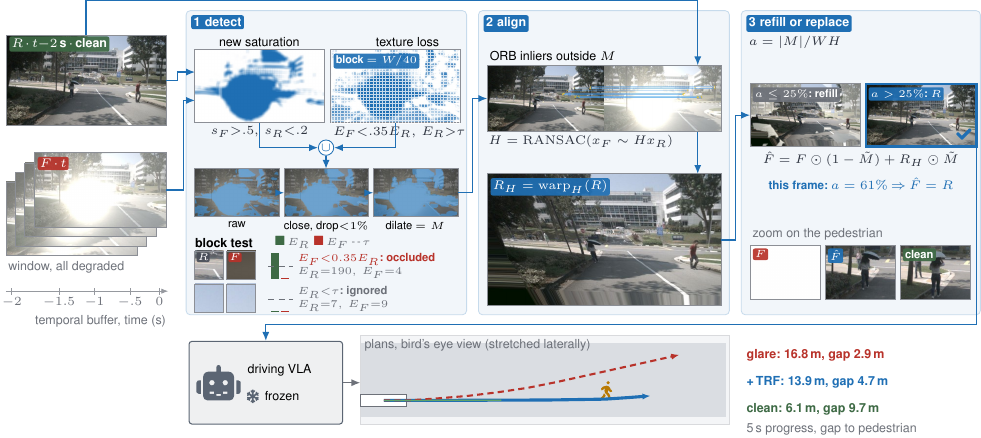}
  \caption{\method{} on the scene of Fig.~\ref{fig:overview}, with
  severity 5 glare over the pedestrian in every frame of the input window.
  \textbf{1} Two cues are computed on blocks of $W/40$ pixels of the current
  frame $F$ against the reference $R$. Their union is closed, small components
  are dropped, and the result is dilated to give $M$. The block test uses the
  occlusion frame of the same scene. An occluded block loses its texture, and a
  sky block is excluded because it had little texture in $R$. \textbf{2} A
  homography is fitted to ORB matches outside $M$ and warps $R$ to the view of
  $F$. \textbf{3} A small region is filled from the warped reference. In this example, $M$
  covers \mArea{}\% of the frame, so the whole reference is passed on and the
  pedestrian reappears. Bottom: plans of the unchanged driving VLA with and
  without the repair.}
  \label{fig:method}
\end{figure*}

\method{} stores the last frame of each camera that is known to be clean, the
reference $R$. When a new frame $F$ arrives, \method{} detects the degraded
region. A small region is replaced by the reference aligned to the view of $F$.
When most of the frame is lost, the whole frame is replaced by the reference.
The unchanged planner receives the repaired frame. Under a persistent degradation every frame in the input window is
degraded, so $R$ is the frame before the window, 2\,s before the current frame.
Fig.~\ref{fig:method} shows each step on a frame with glare over
the hazard, together with the resulting plans.

\paragraph{Degradation detection}
Detection uses two physical cues, both measured relative to $R$ on blocks of
about $1/40$ of the image width. The block scale absorbs the change in content
caused by ego motion between $R$ and $F$. The first cue is new saturation. A
block is flagged when more than half of its pixels are clipped in $F$ and fewer
than a fifth are clipped in $R$. The second cue is texture loss. Let $E_F$ and
$E_R$ be the block mean of the gradient magnitude. A block is flagged when
\begin{equation}
  E_F < 0.35\,E_R \quad\text{and}\quad E_R > \tau ,
\end{equation}
where $\tau$ is the 30th percentile of $E_R$ over the image. The threshold
$\tau$ excludes regions such as sky and road that are already flat in $R$.
The flagged blocks are closed morphologically, components below 1\% of the
image are removed, and the mask $M$ is dilated by one block.

\paragraph{Reference alignment}
Ego motion between $R$ and $F$ causes misalignment between $R$ and $F$.
\method{} detects ORB keypoints \citep{orb} in $F$ outside $M$ and in $R$,
matches the keypoints, and fits a homography with RANSAC \citep{ransac}. The reference
is warped to the view of $F$. A homography models the road plane exactly and
distant structure approximately, and the two make up most of a forward view. If fewer than 25 inlier
matches remain, the unwarped reference is used.

\paragraph{Refill}
The repaired frame is
\begin{equation}
  \hat F = F\odot(1-\tilde M) + \operatorname{warp}_H(R)\odot\tilde M ,
\end{equation}
where $\tilde M$ is $M$ smoothed with a Gaussian of about $1/60$ of the image
width. Pixels outside the degraded region come from the current frame. The
refilled region restores the last observed appearance of the hazard
to the planner input.

\paragraph{Adaptive switch}
Region refill preserves intact regions of the current view. When the degraded
region covers much of the frame, the refilled frame becomes a patchwork of two
viewpoints, with seams at the junction of the two views. \method{} therefore switches by the
flagged area $a = |M| / (WH)$:
\begin{equation}
  \hat F =
  \begin{cases}
    F, & a = 0,\\
    F\odot(1-\tilde M) + \operatorname{warp}_H(R)\odot\tilde M, & 0 < a \le 0.25,\\
    R, & a > 0.25.
  \end{cases}
\end{equation}
The threshold of a quarter of the image is fixed before the evaluation and is not tuned on collisions. The detector runs on every frame of every camera, so
the switch also decides when a frame needs no repair.

\paragraph{\method{}-G}
Two results on the development scenes (Section~\ref{sec:experiments}) lead to
a second form, \method{}-G, which is fixed before evaluation on held-out
scenes. First, the unaligned reference performs better than the aligned one,
so \method{}-G drops the homography. Second, fog degrades every frame and the
reference alike, so no clean reference exists. \method{}-G therefore separates
a global from a local loss. Let $s$ be the share of $M$ that is new
saturation. When $a > 0.5$ and $s < 0.1$, the loss is large and unsaturated,
and \method{}-G restores $F$ with the dark channel prior \citep{dcp}.
Otherwise, \method{}-G applies the adaptive switch with $\operatorname{warp}_H(R)$
replaced by $R$. The thresholds follow from the detector statistics, which
flag 0.67 of the frame under fog and 0.33 under occlusion. The dehazing step is chosen among the dark channel prior, AdaIR \citep{adair} and MoCE-IR
\citep{moceir} on the development scenes under fog only.

\section{Experiments}
\label{sec:experiments}

\subsection{Experimental Setup}
\label{sec:setup}

\paragraph{Planners and scenes}
The main experiments use \model{} \citep{autovla} with the released checkpoint
and the decoding settings of the released evaluation configuration.
The benchmark and the repair are repeated on two further driving VLAs in Section~\ref{sec:planners}. Qwen-Drive-1.0 (4B) reads the same three cameras and four frames
and decodes a 5\,s trajectory with a planning expert \citep{qwendrive}.
Impromptu-VLA (3B) reads the current front frame and a text prompt with the ego
history, and outputs a 3\,s trajectory \citep{impromptuvla}. For Impromptu-VLA
we use the released nuScenes prompts, which cover \nIm{} of our scenes. All
three planners receive the same degraded and repaired frames. The diagnosis covers
\nDiag{} of the 691 scenes. In the remaining three, the decoder stops before ten poses under one condition, so these scenes are left out of the
paired comparisons. The repair experiments cover the \nRep{} scenes that have a clean keyframe 2\,s before the input window.

\paragraph{Metrics}
We report the collision rate against the annotated hazard tracks
(Section~\ref{sec:benchmark}). Every condition is paired with the clean run of
the same scene. A change in collision rate carries a 95\% paired bootstrap
interval, and its significance is an exact McNemar test on the scenes that
change state. Progress over 5\,s and L2 to the human trajectory are reported
as paired differences. \model{} decodes by sampling. Every clean scene is therefore run twice, and the rerun gives the noise floor of each metric. The collision check with oriented boxes and the collisions of the clean plans are reported in Section~\ref{app:collision} of the supplementary material.

\paragraph{Compared repairs}
All compared repairs need no training and leave the planner unchanged. CLAHE
equalises contrast locally on the lightness channel \citep{clahe}. The dark
channel prior removes haze by estimating transmission \citep{dcp}.
Inpainting fills the region detected by \method{} from its surroundings in the
current frame \citep{telea}, so inpainting shares the detector of \method{} but has no
temporal reference. Frame freeze replaces a frame by the reference whenever the
detector flags any region. \method{} is run in two forms. Region refill
replaces only the detected region. The adaptive form replaces the whole frame
when the detected region exceeds a quarter of the image, and refills the region
otherwise. This threshold is fixed before the adaptive runs. All of these
repairs run on every frame of every camera and decide from the detector alone,
as required in deployment. Two privileged references use ground truth about
the degradation: frame freeze applied to the frames known to be degraded, and
\method{} with the true perturbation mask.
Implementation details and all thresholds are listed in
Section~\ref{app:method} of the supplementary material.

\subsection{Diagnosis of Hidden Hazards}
\label{sec:diagnosis}

\begin{table*}[t]
  \centering
  \caption{Safety of \model{} under \bench{} on \nDiag{} in-path hazard scenes.
  Coll.: collision rate (\%). $\Delta$Coll.: change against the clean run with
  a 95\% paired bootstrap interval. New/Res.: scenes that collide only under
  the condition / only in the clean run. $p$: exact McNemar test.
  $\Delta$Prog.: change in 5\,s progress (m). $\Delta$L2: change in L2 (m).}
  \label{tab:diag}
  \small
\begin{tabular}{lrrrrrr}
\toprule
Condition & Coll. & $\Delta$Coll. [95\% CI] & New/Res. & $p$ & $\Delta$Prog. & $\Delta$L2 \\
\midrule
Clean & 16.1 & -- & -- & -- & -- & -- \\
\midrule
Clean, rerun & 16.0 & -0.1 [-0.7, +0.4] & 2/3 & 1.000 & +0.05 & +0.01 \\
\midrule
Glare $s{=}3$, on & 19.9 & +3.8 [+2.0, +5.7] & 34/8 & $<$0.001 & +0.07 & +0.05 \\
Glare $s{=}3$, off & 15.7 & -0.4 [-2.2, +1.5] & 20/23 & 0.761 & +0.15 & +0.02 \\
Glare $s{=}5$, on & 20.9 & +4.8 [+2.8, +7.0] & 44/11 & $<$0.001 & +0.85 & +0.14 \\
Glare $s{=}5$, on, sudden & 16.6 & +0.4 [-1.6, +2.5] & 28/25 & 0.784 & +0.87 & +0.08 \\
Occlusion $s{=}3$, on & 20.9 & +4.8 [+3.1, +6.8] & 39/6 & $<$0.001 & +0.64 & +0.08 \\
Occlusion $s{=}3$, off & 16.9 & +0.7 [-0.9, +2.5] & 19/14 & 0.487 & -0.03 & +0.03 \\
\midrule
Tunnel exit $s{=}5$, sudden & 16.6 & +0.4 [-1.6, +2.5] & 27/24 & 0.780 & +0.69 & +0.09 \\
Fog $s{=}4$ & 20.2 & +4.1 [+1.9, +6.2] & 42/14 & $<$0.001 & +0.07 & +0.11 \\
Rain $s{=}4$ & 16.6 & +0.4 [-1.6, +2.5] & 30/27 & 0.791 & +0.32 & +0.10 \\
\bottomrule
\end{tabular}

\end{table*}

Table~\ref{tab:diag} reports all conditions. Two clean runs of the same scene
differ by \dcollRerun{} points, so the rerun sets the noise floor.

\paragraph{Effect of placement}
For \model{}, concealing the hazard raises the collision rate in every family
that can be placed. Occlusion on the hazard raises the collision rate from \collClean{}\% to
\collOcOn{}\% ($p$\,=\,\pOcOn{}). Glare of severity 3 on the hazard raises
the rate to \collGlThreeOn{}\% ($p$\,=\,\pGlThreeOn{}). The same degradations placed
on the far side of the frame change the rate by \dcollOcOff{} and
\dcollGlThreeOff{} points, within the noise floor. The off placement degrades
the same camera at the same image row and severity. The difference between the
two placements is therefore the effect of losing the hazard. Table~\ref{tab:onoff} tests the on and off placements against each
other directly.

\paragraph{Speed of the failing plans}
A planner that recognises missing information would reduce speed. Among the
scenes that collide only under occlusion on the hazard, \fastOcOn{}\% of plans
cover more distance than the clean plan, by \dprogNewOcOn{}\,m on average.
Under severity 5 glare, the share is \fastGlFiveOn{}\%. The added collisions
therefore come from plans that maintain or increase speed
(Table~\ref{tab:speedup} in the supplementary material).

\paragraph{Effect of clean history}
Severity 5 glare raises the collision rate to \collGlFiveOn{}\% when glare affects
all four input frames. When glare affects only the current frame, the rate is
\collGlFiveSudden{}\% ($p$\,=\,\pGlFiveSudden{}). The tunnel exit, also
sudden, changes the rate by \dcollTunnel{} points. With three clean frames in
the window, the hazard observed earlier still informs the plan. This is the
property that \method{} uses.

\paragraph{Global degradations}
Fog raises the collision rate to \collFog{}\% ($p$\,=\,\pFog{}). Rain changes the rate
by \dcollRain{} points. A breakdown of every condition by scene attribute is given in Section~\ref{app:results} of the supplementary material.

\subsection{Generalization Across Planners}
\label{sec:planners}

\begin{table*}[t]
  \centering
  \caption{\bench{} on three driving VLAs. Each cell: change in collision rate
  (points) / change in progress at the planner's horizon (m), against the clean
  run of the same planner and scene. $^*$: $p<0.05$ (McNemar for collisions,
  Wilcoxon for progress). Clean: collision rate of the clean run (\%).
  Collisions within 3\,s are rare for every planner
  (Section~\ref{app:collision} of the supplementary material), so Impromptu-VLA is compared on progress.}
  \label{tab:models}
  \small
\begin{tabular}{llrrccccc}
\toprule
Planner & Horizon & $n$ & Clean & Occl.\ on & Occl.\ off & Glare$_5$ on & Glare$_3$ off & Fog \\
\midrule
AutoVLA & 5\,s & 688 & 16.1 & +4.8$^*$ / +0.64$^*$ & +0.7 / -0.03 & +4.8$^*$ / +0.85$^*$ & -0.4 / +0.15 & +4.1$^*$ / +0.07 \\
Qwen-Drive & 5\,s & 668 & 2.4 & +0.6 / -1.10$^*$ & -0.1 / +0.02 & +0.7 / -0.59$^*$ & +2.8$^*$ / +0.41$^*$ & +8.7$^*$ / +0.23$^*$ \\
Impromptu-VLA & 3\,s & 134 & 0.0 & +0.0 / +0.62$^*$ & +0.0 / +0.42 & +0.0 / +0.38$^*$ & +0.0 / +0.06 & +0.0 / +0.63$^*$ \\
\bottomrule
\end{tabular}

\end{table*}

Table~\ref{tab:models} applies \bench{} to all three planners. The planners
respond differently to a concealed hazard. The on minus off comparison
isolates that response, since both placements degrade the image equally.
\model{} covers \onoffAvOc{}\,m more under occlusion on the hazard than away from the hazard
($p$\,=\,\onoffPAvOc{}). Impromptu-VLA covers \onoffImOc{}\,m more over its
3\,s horizon ($p$\,=\,\onoffPImOc{}). Both maintain or increase speed when the
hazard is concealed. Qwen-Drive instead reduces speed. The Qwen-Drive plan covers \onoffQdOc{}\,m under
occlusion on the hazard ($p$\,=\,\onoffPQdOc{}) and \onoffQdGl{}\,m under glare
on the hazard ($p$\,=\,\onoffPQdGl{}). Qwen-Drive reduces speed when part of the scene is
lost, with no increase in collision rate under local degradation
(\diagQdOc{} points under occlusion, $p$\,=\,\diagPQdOc{}).

Under fog, the collision rate of Qwen-Drive rises by \diagQdFog{}
points ($p$\,=\,\diagPQdFog{}), the largest effect of any condition on any
planner. \bench{} therefore separates two failure modes that a whole-frame
benchmark reports as one number. In the first, the plan maintains speed when
the hazard is concealed. In the second, the plan reduces speed under a local loss
but maintains speed under a global loss of contrast.

Table~\ref{tab:modelsrepair} applies adaptive \method{} unchanged to each
planner. Adaptive \method{} lowers the collision rate under the conditions that degrade each planner. For \model{}
this is occlusion, glare and fog. For Qwen-Drive the relevant condition is fog, from
\repQdFogBefore{}\% to \repQdFogAfter{}\% ($p$\,=\,\repQdFogP{}). Under
occlusion and glare, Qwen-Drive already reduces speed, and the repair restores the usual Qwen-Drive speed. The Qwen-Drive collision rate changes from \repQdOcBefore{}\% to
\repQdOcAfter{}\% and from \repQdGlBefore{}\% to \repQdGlAfter{}\%, neither
change significant. On clean input the rate stays at \repQdClAfter{}\% against
\repQdClBefore{}\% ($p$\,=\,\repQdClP{}).

\subsection{Validation on Real Rain and Night Scenes}
\label{sec:weather}

We repeat the core test on real recordings, where the adverse
condition also changes the scene. Using the nuScenes scene descriptions, we add
every in-path hazard scene from recordings labelled as rain or night that
passes the selection of Section~\ref{sec:benchmark}. This adds
\wNAvRain{} rain scenes, \wNAvNight{} night scenes and further day scenes from
the same recordings. On these scenes we run the clean input, occlusion on and
off the hazard, and adaptive \method{} on the clean and the occluded input, for
\model{} and Qwen-Drive.

\begin{table*}[t]
  \centering
  \caption{Real rain and night scenes. Clean: collision rate of the clean run
  (\%). Overshoot: plan progress minus human progress over the horizon (m).
  Occl.\ on$-$off: change in collision rate (points) and in progress (m) when
  the same occlusion covers the hazard instead of the far side of the frame.
  \method{}: collision rate (\%) without $\to$ with adaptive \method{}, on the
  clean input and under occlusion on the hazard. $^*$: $p<0.05$ (McNemar for
  collisions, Wilcoxon for progress).}
  \label{tab:weather}
  \small
\begin{tabular}{llrrrrrcc}
\toprule
Planner & Scenes & $n$ & Clean & Overshoot & Occl.\ on$-$off & $\Delta$Prog.\ on$-$off & \method{}, clean & \method{}, occl.\ on \\
\midrule
AutoVLA & day & 842 & 15.7 & +2.67 & +4.8$^*$ & +0.80$^*$ & 15.6 $\to$ 14.7 & 21.2 $\to$ 16.8$^*$ \\
AutoVLA & rain & 186 & 35.5 & +2.78 & +2.2 & +1.10$^*$ & 33.5 $\to$ 33.5 & 36.0 $\to$ 34.2 \\
AutoVLA & night & 1040 & 18.3 & -3.52 & +4.4$^*$ & +2.75$^*$ & 17.8 $\to$ 18.5 & 22.9 $\to$ 19.2$^*$ \\
\midrule
Qwen-Drive & day & 817 & 3.5 & +0.29 & +1.7$^*$ & -1.06$^*$ & 3.9 $\to$ 3.8 & 5.6 $\to$ 6.9 \\
Qwen-Drive & rain & 179 & 1.1 & +0.29 & -0.6 & -1.84$^*$ & 1.2 $\to$ 0.6 & 0.0 $\to$ 2.5 \\
Qwen-Drive & night & 1004 & 5.1 & -2.17 & +9.0$^*$ & -0.51$^*$ & 5.7 $\to$ 5.5 & 13.9 $\to$ 7.0$^*$ \\
\bottomrule
\end{tabular}

\end{table*}

Table~\ref{tab:weather} reports the results. The effect of concealing the hazard
persists under real conditions. At night, \model{} covers \wOnOffProgAvNight{}\,m
more when the occlusion covers the hazard, the largest change in
Table~\ref{tab:weather}. The AutoVLA collision rate rises by \wOnOffAvNight{} points
($p$\,=\,\wOnOffPAvNight{}). Qwen-Drive still reduces progress at night, by
\wOnOffProgQdNight{}\,m, about half as much as in the day. The Qwen-Drive collision rate
rises by \wOnOffQdNight{} points ($p$\,=\,\wOnOffPQdNight{}). In rain, both planners
change their progress significantly, while the change in collisions does not
reach significance on \wNAvRain{} scenes.

Adaptive \method{} also reduces collisions in these conditions. Under occlusion at night, adaptive \method{} lowers the
collision rate of \model{} from \wTrfBeforeAvNight{}\% to \wTrfAfterAvNight{}\%
($p$\,=\,\wTrfPAvNight{}). For Qwen-Drive the rate falls from
\wTrfBeforeQdNight{}\% to \wTrfAfterQdNight{}\% ($p$\,=\,\wTrfPQdNight{}). On clean rain and night input,
where the whole view is darker or streaked, adaptive \method{} leaves both collision rates statistically unchanged.

\begin{figure*}[t]
  \centering
  \includegraphics{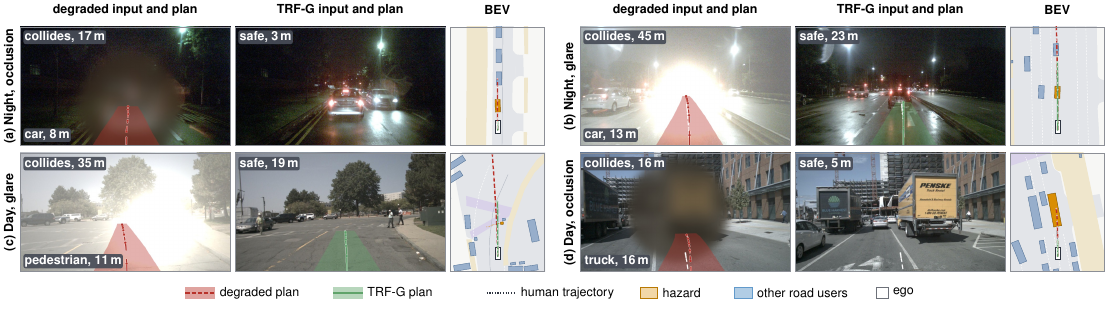}
  \caption{Held-out scenes in which the clean plan of \model{} is
  collision-free, the plan under degradation collides, and the plan after
  \method{}-G is collision-free. Tags give the 5\,s progress of each plan and the
  hazard distance. The plan is drawn on the road as a band of the ego width.
  BEV: bird's eye view with the nuScenes map, every annotated road user and
  the human trajectory.}
  \label{fig:examples}
\end{figure*}

\subsection{Repair on the Development Scenes}
\label{sec:compare}

Table~\ref{tab:compare} in the supplementary material compares the repairs
on the development scenes. Adaptive \method{} lowers the collision rate under
occlusion on the hazard from \repOcOnNone{}\% to \repOcOnAdapt{}\%
($p$\,=\,\repPOcOnAdapt{}). Under severity 5 glare, the collision rate falls from \repGlFiveOnNone{}\%
to \repGlFiveOnAdapt{}\% ($p$\,=\,\repPGlFiveOnAdapt{}). CLAHE, the dark channel
prior and inpainting give \repOcOnClahe{}\%, \repOcOnDcp{}\% and
\repOcOnInpaint{}\% under occlusion, with no significant change. Inside an occluded or
saturated region the current frame holds no trace of the hazard, and a temporal
reference is the only source that still contains the hazard. Under fog the dark channel
prior gives \repFogDcp{}\% and adaptive \method{} \repFogAdapt{}\%, since fog
leaves a weak trace of every object in the current frame.

\subsection{Held-Out Comparison with Recent Methods}
\label{sec:heldout}

\begin{table*}[t]
  \centering
  \caption{Held-out comparison on \hoAvNGl{} to \hoAvNCl{} scenes per condition for
  \model{} and \hoQdNOc{} for Qwen-Drive, none of them used to design \method{}.
  Collision rate (\%). Bold: lowest repair in each column. $^*$/$^\times$:
  lower/higher than no repair, $p<0.05$. $^\dagger$: differs from
  \method{}-G, $p<0.05$. All tests are exact McNemar tests. Every repair runs on every frame of every
  camera. ProPainter receives the reference frame, the four input frames and the
  mask of \method{}. Avg.: mean over the eight columns. ms: time per
  $1600\times900$ frame, on one CPU core for the training-free repairs and on one
  A100 GPU for the learned models.}
  \label{tab:heldout}
  \resizebox{\textwidth}{!}{
\begin{tabular}{lcccccccc|cc}
\toprule
 & \multicolumn{4}{c}{AutoVLA} & \multicolumn{4}{c}{Qwen-Drive} & & \\
\cmidrule(lr){2-5}\cmidrule(lr){6-9}
Repair & Clean & Occl.\ on & Glare$_5$ on & Fog & Clean & Occl.\ on & Glare$_5$ on & Fog & Avg. & ms \\
\midrule
No repair & 19.8 & 25.6 & 25.3 & 20.8 & 5.2 & 13.4 & 11.5 & 5.2 & 15.8 & -- \\
CLAHE \citep{clahe} & 21.3 & 25.2$^\dagger$ & 25.3 & 22.0 & 4.5 & 10.9$^*$$^\dagger$ & 10.7$^\dagger$ & 7.5$^\times$$^\dagger$ & 15.9 & 29 \\
Dark channel prior \citep{dcp} & \textbf{19.3} & 25.4$^\dagger$ & 24.8 & 21.2 & \textbf{3.5}$^*$ & 11.5$^*$$^\dagger$ & 8.7$^*$ & 4.5 & 14.9 & 246 \\
Inpainting \citep{telea} & 21.0 & 27.3$^\dagger$ & 25.5$^\dagger$ & 23.2 & 4.5 & 10.0$^*$$^\dagger$ & 10.4$^\dagger$ & 9.3$^\times$$^\dagger$ & 16.4 & 183 \\
AdaIR \citep{adair} & 20.3 & 26.1$^\dagger$ & 25.1 & 20.3 & 4.5 & 11.5$^*$$^\dagger$ & 10.5$^\dagger$ & 4.8 & 15.4 & 214 \\
MoCE-IR \citep{moceir} & 20.0 & 26.2$^\dagger$ & \textbf{23.2}$^*$ & 21.0 & 4.2 & 12.5$^\dagger$ & 10.9$^\dagger$ & \textbf{4.0} & 15.3 & 194 \\
Flare7K++ \citep{flare7kpp} & 20.7 & 26.9$^\dagger$ & 25.5 & \textbf{19.7} & 4.3 & 13.2$^\dagger$ & 12.2$^\dagger$ & 5.5 & 16.0 & 185 \\
ProPainter \citep{propainter} & 20.8 & 24.2$^\dagger$ & 23.9 & 22.9$^\dagger$ & 4.5 & 8.3$^*$ & 8.7$^*$ & 10.0$^\times$$^\dagger$ & 15.4 & 209 \\
\midrule
Frame freeze & 21.2 & 23.3$^*$$^\dagger$ & 23.6 & 22.9$^\dagger$ & 6.2$^\dagger$ & \textbf{7.0}$^*$ & \textbf{7.0}$^*$ & 7.2$^\dagger$ & 14.8 & 68 \\
\method{} (adaptive) & 20.5 & 22.7$^*$$^\dagger$ & 23.8 & 22.7 & 4.0$^*$ & 7.3$^*$ & 7.5$^*$ & 7.2$^\dagger$ & 14.5 & 108 \\
\method{}-G & 20.7 & \textbf{21.5}$^*$ & 23.4 & 20.5 & 4.3 & 7.3$^*$ & 7.7$^*$ & 4.2 & \textbf{13.7} & 127 \\
\bottomrule
\end{tabular}
}
\end{table*}

\method{}-G, the compared methods, the split and the tests are fixed in a written protocol before any held-out run. The 599 held-out scenes are drawn
with a fixed seed from the rain, night and day recordings of
Section~\ref{sec:weather}. Four recent learned models join the repairs above,
with their released weights: AdaIR \citep{adair} and MoCE-IR \citep{moceir}
for all-in-one restoration, Flare7K++ \citep{flare7kpp} for flare removal,
and ProPainter \citep{propainter} for video inpainting. The all-in-one networks
run at 800$\times$450 pixels, since at full resolution these networks leave the frames
nearly unchanged.

Table~\ref{tab:heldout} reports all methods and conditions. Under occlusion of
the hazard, \method{}-G gives the lowest collision rate for \model{}, from
\hoAvOcNone{}\% to \hoAvOcG{}\% ($p$\,=\,\hoAvOcGP{}). This rate is lower than that of every
other repair with $p<0.05$, including ProPainter (\hoAvOcPro{}\%) and adaptive
\method{} (\hoAvOcAdapt{}\%). For Qwen-Drive, \method{}-G lowers the rate from
\hoQdOcNone{}\% to \hoQdOcG{}\%, while the learned restoration models give
\hoQdOcAdair{}\% to \hoQdOcFlare{}\%. Under glare on the hazard, \method{}-G
lowers the rate of Qwen-Drive from \hoQdGlNone{}\% to \hoQdGlG{}\%
($p$\,=\,\hoQdGlGP{}). For \model{}, \method{}-G gives \hoAvGlG{}\%, level with MoCE-IR
(\hoAvGlMoce{}\%). Flare7K++ does not lower the rate for either planner.

Averaged over both planners and all four conditions, \method{}-G gives
\hoAvgG{}\% against \hoAvgNone{}\% without repair. The best learned model,
MoCE-IR, gives \hoAvgMoce{}\%, and the best repair from prior work, the dark
channel prior, gives \hoAvgDcp{}\%. \method{}-G runs in \hoMsG{}\,ms per frame on one CPU core. The
learned models need \hoMsMoce{}\,ms to \hoMsAdair{}\,ms per frame.

Frame freeze triggered by the detector is close to \method{}-G under local
degradation for Qwen-Drive. Detector-triggered frame freeze raises the rate on clean input to
\hoQdClFreeze{}\% against \hoQdClG{}\% for \method{}-G ($p$\,=\,\hoQdClFreezePG{}),
and under fog to \hoQdFogFreeze{}\%. \method{}-G maintains the rate under fog at
\hoQdFogG{}\% because \method{}-G routes fog to dehazing, where adaptive \method{} and
frame freeze pass a reference that is itself foggy. On the held-out scenes, fog raises the rate
of neither planner significantly, so no repair has an effect to recover there.

\subsection{Ablation Study}
\label{sec:ablation}

Table~\ref{tab:ablation} in the supplementary material removes one component
of region refill at a time on the development scenes. The saturation cue alone
helps under glare and the texture cue alone under occlusion and fog. Removing
the homography lowers the collision rate under occlusion from \repOcOnTrf{}\%
to \repOcOnNoAlign{}\% and under glare from \repGlFiveOnTrf{}\% to
\repGlFiveOnNoAlign{}\%. Over 2\,s the homography stretches near objects and
the image border, and these artefacts cost the planner more than the age of the
reference. \method{}-G therefore omits the homography.

\subsection{Cost on Clean Input}
\label{sec:cleancost}

A repair runs on every frame and must preserve performance on clean input. On the
development scenes, frame freeze triggered by the detector is the only repair
that raises the clean collision rate, from \cleanNone{}\% to
\cleanDetFreeze{}\%. The detector flags some area on most clean frames, and
frame freeze replaces every flagged frame.
Adaptive \method{} gives \cleanAdapt{}\% (Table~\ref{tab:cleancost} in the
supplementary material). The held-out scenes show the same pattern for
Qwen-Drive (Table~\ref{tab:heldout}).

\subsection{Qualitative Results}

Fig.~\ref{fig:examples} shows four held-out scenes, two of them at night. In
each, the degraded plan intersects the concealed hazard. After \method{}-G the
hazard is visible again, and the plan stops behind the hazard or maintains clearance. In (a)
the occluded frame is largely lost, so \method{}-G passes the reference, which
shows the car ahead.

\section{Conclusion}

This paper introduced \bench{}, a camera degradation benchmark for end-to-end
driving that indexes the degradation to the in-path hazard. Matched on and off
placements separate the loss of the hazard from the loss of image quality. On
three driving VLAs, concealing the hazard leads two planners to maintain or
increase speed and Qwen-Drive to reduce speed. Qwen-Drive shows a higher
collision rate under fog. Sudden degradation with clean frames in the input
history has little effect on safety. Building on this finding, \method{}
repairs the input at test time from the last clean frame. \method{} refills a
small degraded region and replaces a largely lost frame with the reference.
\method{} lowers the collision rate under the conditions that degrade each
planner, with no significant increase on clean input. On held-out scenes, the
variant \method{}-G, which routes fog to dehazing, achieves a lower collision
rate under hazard occlusion than recent learned restoration and video
inpainting models.

Future work will extend \bench{} to further VLA and modular planners and to
closed-loop simulation. Future work will also add LiDAR depth to the
alignment, so that near hazards are refilled at their current position despite
parallax.

\clearpage
\appendices
\renewcommand{\thesection}{S\arabic{section}}
\renewcommand{\thetable}{S\arabic{table}}
\renewcommand{\thefigure}{S\arabic{figure}}
\renewcommand{\thealgorithm}{S\arabic{algorithm}}
\renewcommand{\theequation}{S\arabic{equation}}
\setcounter{section}{0}\setcounter{table}{0}\setcounter{figure}{0}\setcounter{equation}{0}
\section{Dataset Details}
\label{app:dataset}

\subsection{Scene Selection}
\bench{} starts from the nuScenes keyframes for which the ego vehicle has a
complete 5\,s future, 5{,}043 keyframes in total. Every annotation is
transformed into the ego frame of the front camera keyframe, with $x$
forward and $y$ to the left. An annotation is an in-corridor hazard when the
category is a vehicle, a pedestrian or a cyclist. The annotation centre must also satisfy
$0 < x \le 35$\,m and $|y| \le 2$\,m. This rule retains 1{,}151 keyframes.

The corridor follows the ego heading, so the corridor admits agents on curved roads
that the vehicle never approaches. We therefore densify the human trajectory
over the next 5\,s and keep a scene only when the nearest hazard lies within
2.5\,m of the trajectory. When the vehicle stops before reaching the hazard, the lateral
offset of the hazard from the path, interpolated at the hazard's distance,
must be below 1.5\,m. The hazard centre must also project into the front
camera. This leaves 691 scenes. Table~\ref{tab:dataset} summarises these scenes.

\begin{table}[h]
  \centering
  \caption{Statistics of the 691 \bench{} scenes.}
  \label{tab:dataset}
  \resizebox{\columnwidth}{!}{
\begin{tabular}{lr}
\toprule
Statistic & Value \\
\midrule
Scenes (keyframes) & 691 \\
Drives & 80 \\
Singapore / Boston & 426 / 265 \\
With $t{-}2$\,s reference & 600 \\
Hazards in corridor per scene, mean / max & 1.44 / 13 \\
Nearest hazard distance, median [IQR] (m) & 18.8 [10.8, 26.9] \\
Ego speed, median (m/s) & 1.90 \\
Ego stopped ($<$0.5\,m/s) & 247 \\
Command keep / left / right & 636 / 15 / 40 \\
Nearest hazard: vehicle.car & 429 \\
Nearest hazard: human.pedestrian.adult & 141 \\
Nearest hazard: human.pedestrian.construction worker & 29 \\
Nearest hazard: vehicle.bus.rigid & 29 \\
Nearest hazard: vehicle.truck & 28 \\
Nearest hazard: vehicle.motorcycle & 14 \\
Nearest hazard: vehicle.bicycle & 14 \\
Nearest hazard: vehicle.trailer & 7 \\
\bottomrule
\end{tabular}
}
\end{table}

\subsection{Hazard Projection}
Let camera $c$ have intrinsics $K_c$ and extrinsics $(R_c, t_c)$ relative
to the ego frame. The hazard centre $p$ is mapped to
\begin{align}
  (u, v, w)^\top &= K_c\, R_c^\top (p - t_c), \\
  (\hat u, \hat v) &= \bigl(u/(wW),\; v/(wH)\bigr),
\end{align}
with $W\times H = 1600\times 900$. Each camera has its own timestamp. The point $p$ is therefore first moved to the world frame with the keyframe ego pose and then to
the ego pose of that camera image. A hazard is visible in camera $c$ when its
depth exceeds 0.5\,m and $(\hat u, \hat v) \in [0,1]^2$.

\subsection{Degradation Models}
Table~\ref{tab:params} lists every degradation model and how its parameters
depend on the severity $s \in \{1,\dots,5\}$. Write $\sigma = s/5$ and
let $r$ be the distance to the source normalised by the longer image side.

\begin{table*}[t]
  \centering
  \caption{Degradation models of \bench{}. $x$: image in $[0,1]$.
  Random quantities are drawn from a generator seeded by the image name.
  Every degraded frame can therefore be reproduced.}
  \label{tab:params}
  \small
  \begin{tabular}{lp{0.70\textwidth}}
    \toprule
    Family & Model and parameters \\
    \midrule
    Glare & Core $2.5\sigma\exp(-(r/(0.03+0.05\sigma))^2)$, bloom
      $1.2\sigma/(1+(r/(0.04+0.10\sigma))^2)$, tinted $(1, 0.95, 0.85)$;
      veil $x(1-0.35\sigma)+0.35\sigma$; three ghosts at $-0.6, -1.2, 0.5$ times
      the source offset through the image centre, amplitudes $0.10, 0.06, 0.05$;
      clipping to $[0,1]$. Source at the projected hazard (on) or at
      $(0.15$ or $0.85, \hat v)$ (off). \\
    Tunnel exit & Tone curve $(1-e^{-gx})/(1-e^{-g})$ with $g=1+4\sigma$;
      opening term $0.8\sigma$ times a Gaussian centred at $(0.5W, 0.3H)$ with
      widths $(0.45W, 0.35H)$; veil $0.3\sigma$. \\
    Occlusion & Covered fraction $0.08+0.13(s-1)$. Placed: one disc of that
      area at the target point. Scattered: random ellipses until the fraction
      is reached. Mask blurred with $\sigma_b = 0.02W$; content inside replaced by
      $0.35$ times a blur of the image plus $0.65$ times a dirt colour
      $(0.20, 0.17, 0.12)$. \\
    Fog & $I = J\,t + A(1-t)$, $t=\exp(-\beta d)$, $\beta = 0.3+4\sigma$,
      $A=0.82$. Depth proxy per row $y\in[0,1]$:
      $d=\min(0.08/(\max(y-0.45,0)+0.08),\,0.7)$. \\
    Rain & $1500\sigma$ drops per $1600\times900$ pixels, streak length
      $15+40\sigma$\,px at $15^\circ$; veil $0.25\sigma$; Gaussian blur
      $0.5+1.5\sigma$\,px. \\
    \bottomrule
  \end{tabular}
\end{table*}

\begin{figure*}[t]
  \centering
  \includegraphics{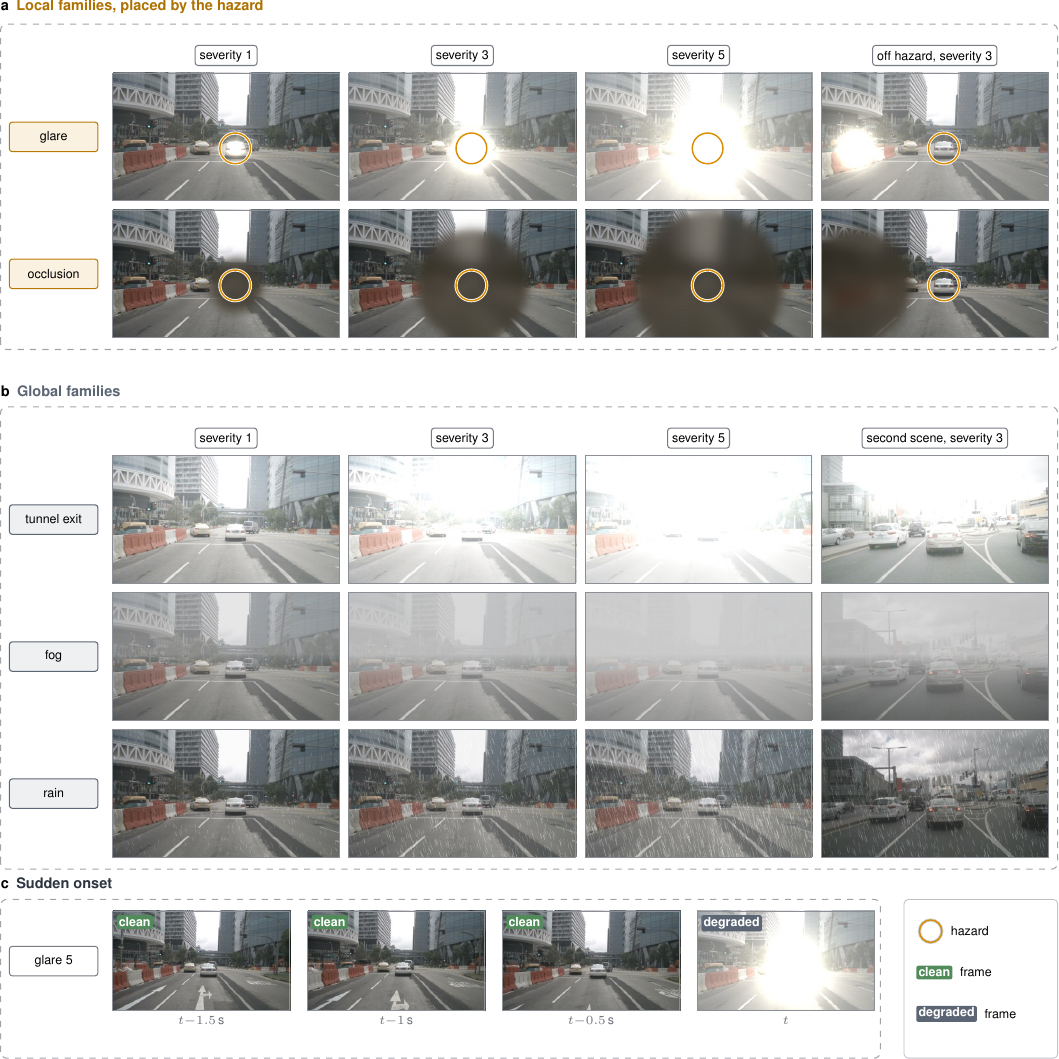}
  \caption{\bench{} degradations on one front camera frame with a car 18.4\,m
  ahead. The ring marks the projected hazard. \textbf{(a)} Glare and occlusion
  at severities 1, 3 and 5 are centred on the hazard. In the last column, the
  same degradation at severity 3 is placed on the far side of the frame.
  \textbf{(b)} The global families at the same severities, with a second scene
  in the last column. \textbf{(c)} The four input frames under sudden onset.
  Only the current frame is degraded. A persistent onset degrades all four
  frames.}
  \label{fig:gallery}
\end{figure*}

\subsection{Conditions and Counts}
Each scene is run under the eleven conditions of Table~\ref{tab:diag}. Glare
and occlusion are applied to the front camera only, where the hazard is
visible. Tunnel exit, fog and rain are applied to all three front cameras.
A persistent condition degrades the four frames of the input window and a
sudden condition degrades only the current frame. Per scene this gives 48
degraded input frames, so the diagnosis uses 33{,}168 degraded frames.

\section{Method Details}
\label{app:method}

\subsection{Algorithm}
Algorithm~\ref{alg:trf} gives \method{} for one camera. In deployment the
reference can be updated whenever a frame passes the detector with an empty
mask, to retain the most recent clean frame. In our evaluation every frame of
the input window can be degraded, and the reference is the keyframe 2\,s
before the current frame.

\begin{algorithm}[h]
\caption{\method{} for one camera stream}
\label{alg:trf}
\begin{algorithmic}[1]
\Require frame $F_t$, reference $R$ (last clean frame)
\State $k \gets \max(9, \lfloor W/40 \rfloor)$, odd
\State $S_F, S_R \gets$ block means of $[\,\mathrm{gray} > 0.97\,]$ in $F_t$, $R$
\State $E_F, E_R \gets$ block means of $\lVert\nabla\,\mathrm{gray}\rVert$ in $F_t$, $R$
\State $\tau \gets$ 30th percentile of $E_R$
\State $M \gets (S_F > 0.5 \wedge S_R < 0.2) \vee (E_F < 0.35E_R \wedge E_R > \tau)$
\State $M \gets \mathrm{dilate}_k(\mathrm{drop}_{<1\%}(\mathrm{close}_k(M)))$
\If{$M = \emptyset$}
  \State $R \gets F_t$; \Return $F_t$
\EndIf
\State match ORB keypoints of $R$ to those of $F_t$ outside $M$
\State $H \gets \mathrm{RANSAC}$ homography, reprojection threshold 4\,px
\If{fewer than 25 inliers} $H \gets I$ \EndIf
\If{$|M|/(WH) > 0.25$} \Return $R$ \EndIf
\State $\tilde M \gets$ Gaussian blur of $M$, kernel $\max(9, \lfloor W/60\rfloor)$
\State \Return $F_t \odot (1-\tilde M) + \mathrm{warp}_H(R)\odot \tilde M$
\end{algorithmic}
\end{algorithm}

\subsection{Hyperparameters}
Table~\ref{tab:hyper} lists every parameter. All values are fixed before any planner run, and none is tuned on the planner's collisions. Table~\ref{tab:sens}
varies each detection parameter around the default value.

\begin{table}[h]
  \centering
  \caption{Parameters of \method{}. $W$: image width.}
  \label{tab:hyper}
  \resizebox{\columnwidth}{!}{%
  \begin{tabular}{llr}
    \toprule
    Stage & Parameter & Value \\
    \midrule
    Detect & block size $k$ & $W/40$ (41\,px) \\
           & saturation level & 0.97 \\
           & saturated share in $F$ / in $R$ & $>0.5$ / $<0.2$ \\
           & texture ratio $E_F/E_R$ & $<0.35$ \\
           & texture floor $\tau$ & 30th pct.\ of $E_R$ \\
           & minimum component area & 1\% of image \\
    \midrule
    Align  & ORB keypoints & 3{,}000 \\
           & matcher & Hamming, cross-check \\
           & RANSAC threshold & 4\,px \\
           & minimum inliers & 25 \\
    \midrule
    Switch & whole frame above flagged area & 25\% \\
    \midrule
    Refill & feather kernel & $W/60$ (27\,px) \\
           & reference age & 2\,s (keyframe before window) \\
    \bottomrule
  \end{tabular}}
\end{table}

\begin{table}[h]
  \centering
  \caption{Detection sensitivity on 200 scenes. One parameter varies at a time;
  bold marks the default. IoU against the true perturbation mask. Clean flag:
  mean flagged area on clean input.}
  \label{tab:sens}
  \resizebox{\columnwidth}{!}{
\begin{tabular}{lrrrrr}
\toprule
Parameter & Value & IoU occl. & IoU glare & IoU fog & Clean flag (\%) \\
\midrule
Texture ratio & 0.2 & 0.63 & 0.39 & 0.54 & 2.3 \\
Texture ratio & 0.25 & 0.63 & 0.40 & 0.59 & 3.2 \\
Texture ratio & \textbf{0.35} & 0.61 & 0.46 & 0.66 & 5.7 \\
Texture ratio & 0.5 & 0.56 & 0.58 & 0.74 & 12.3 \\
Texture ratio & 0.65 & 0.50 & 0.81 & 0.81 & 22.7 \\
\midrule
Saturated share & 0.3 & 0.60 & 0.49 & 0.66 & 6.2 \\
Saturated share & \textbf{0.5} & 0.61 & 0.46 & 0.66 & 5.7 \\
Saturated share & 0.7 & 0.61 & 0.44 & 0.66 & 5.5 \\
\midrule
Block = $W/$ & 20 & 0.62 & 0.48 & 0.69 & 5.3 \\
Block = $W/$ & \textbf{40} & 0.61 & 0.46 & 0.66 & 5.7 \\
Block = $W/$ & 80 & 0.59 & 0.44 & 0.63 & 5.7 \\
\bottomrule
\end{tabular}
}
\end{table}

\subsection{Detection Quality and Cost}
Table~\ref{tab:detection} measures the detector against the true perturbation
mask on every repair scene, for the full detector and for each cue alone. The
true mask of glare and fog covers most of the image because the veil changes
every pixel slightly. The detector targets the region where content is lost,
which is smaller. IoU against the true mask is therefore a conservative
measure for these families. Rain streaks leave local texture intact, and the
detector does not flag rain. This matches the diagnosis, where rain does not
change the collision rate significantly (Table~\ref{tab:diag}). Table~\ref{tab:detmisc} reports the flagged area on
clean input and the share of frames where the homography is accepted. The table also
reports the reconstruction error inside the occluded region and the runtime of
each stage on one CPU core.

\begin{table}[h]
  \centering
  \caption{Detection quality per family on the repair scenes. True area: mean
  share of the image changed by the degradation (\%).}
  \label{tab:detection}
  \resizebox{\columnwidth}{!}{
\begin{tabular}{lrrrrrrr}
\toprule
 & & \multicolumn{2}{c}{both cues} & \multicolumn{2}{c}{saturation} & \multicolumn{2}{c}{texture} \\
\cmidrule(lr){3-4}\cmidrule(lr){5-6}\cmidrule(lr){7-8}
Family & True area & IoU & Rec. & IoU & Rec. & IoU & Rec. \\
\midrule
Glare $s{=}3$ & 95.6 & 0.19 & 0.19 & 0.10 & 0.11 & 0.15 & 0.15 \\
Glare $s{=}5$ & 97.2 & 0.46 & 0.47 & 0.36 & 0.36 & 0.38 & 0.38 \\
Occlusion $s{=}3$ & 28.3 & 0.60 & 0.82 & 0.00 & 0.00 & 0.60 & 0.82 \\
Tunnel exit $s{=}5$ & 97.3 & 0.78 & 0.79 & 0.74 & 0.75 & 0.64 & 0.65 \\
Fog $s{=}4$ & 95.2 & 0.66 & 0.68 & 0.00 & 0.00 & 0.66 & 0.68 \\
Rain $s{=}4$ & 58.2 & 0.00 & 0.00 & 0.00 & 0.00 & 0.00 & 0.00 \\
\bottomrule
\end{tabular}
}
\end{table}

\begin{table}[h]
  \centering
  \caption{Clean input, alignment, reconstruction error (mean absolute error in
  8-bit units inside the true occluded region) and runtime of \method{}.}
  \label{tab:detmisc}
  \resizebox{0.85\columnwidth}{!}{
\begin{tabular}{lr}
\toprule
Quantity & Value \\
\midrule
Flagged area on clean input, mean & 5.5\,\% \\
Clean frames with flagged area $>$2\,\% & 60.0\,\% \\
Homography accepted (occlusion) & 72.8\,\% \\
Error in true region, no repair & 50.0 \\
Error in true region, unaligned refill & 27.7 \\
Error in true region, aligned refill & 27.6 \\
Detection time per frame (median) & 92\,ms \\
Alignment time per frame (median) & 160\,ms \\
Refill time per frame (median) & 69\,ms \\
\bottomrule
\end{tabular}
}
\end{table}

\subsection{Design Choices}
\paragraph{Temporal reference}
Inside an occluded or saturated region the current frame carries no
information about the scene. Any operator that uses only the current frame,
from contrast equalisation to inpainting, can at best produce a plausible
surface. The temporal reference is the only source in a camera-only system
that contains the hazard.

\paragraph{Block statistics}
Ego motion between the reference and the current frame changes the content
of individual pixels. Comparing block means of saturation and gradient energy
measures whether a region still carries texture. Block means therefore tolerate the small shift in content caused by ego motion. As shown in Table~\ref{tab:sens}, halving or doubling the block size moves IoU less than the choice of cue does.

\paragraph{Alignment and the adaptive switch}
A homography aligns the road plane and distant structure, which make up most
of a forward view. The homography leaves near-object parallax unmodelled and
stretches the image border beyond the reference coverage. The ablation in
Table~\ref{tab:ablation} shows that these artefacts have a greater effect on collision rate than the 2\,s age
of the reference. Refilling without alignment, or replacing the frame outright,
gives lower collision rates than aligned refill. The adaptive switch retains the
aligned refill only where the lost region is small. There the seam is short
and the current view still dominates the frame.

\section{Additional Results}
\label{app:results}

\subsection{All Metrics per Condition}
Table~\ref{tab:full} reports the absolute value of every metric for every
condition on the diagnosis scenes. The median gap is the median over scenes of
the minimum distance between the plan and the hazard track. Gap below 2\,m
counts near misses together with collisions.

\begin{table}[h]
  \centering
  \caption{Absolute metrics per condition on the \nDiag{} diagnosis scenes.}
  \label{tab:full}
  \resizebox{\columnwidth}{!}{
\begin{tabular}{lrrrrr}
\toprule
Condition & Coll. & Gap (med.) & Gap$<$2\,m & Prog.\ 5\,s & L2 \\
\midrule
Clean & 16.1 & 10.17 & 10.9 & 16.85 & 1.10 \\
Clean, rerun & 16.0 & 10.10 & 11.0 & 16.90 & 1.11 \\
Glare $s{=}3$, on & 19.9 & 10.07 & 14.5 & 16.92 & 1.15 \\
Glare $s{=}3$, off & 15.7 & 10.07 & 10.6 & 17.00 & 1.12 \\
Glare $s{=}5$, on & 20.9 & 9.65 & 18.5 & 17.70 & 1.24 \\
Glare $s{=}5$, on, sudden & 16.6 & 9.43 & 14.4 & 17.72 & 1.18 \\
Occlusion $s{=}3$, on & 20.9 & 9.67 & 16.3 & 17.49 & 1.18 \\
Occlusion $s{=}3$, off & 16.9 & 10.09 & 11.2 & 16.82 & 1.13 \\
Tunnel exit $s{=}5$, sudden & 16.6 & 10.07 & 11.2 & 17.54 & 1.19 \\
Fog $s{=}4$ & 20.2 & 10.34 & 17.3 & 16.92 & 1.21 \\
Rain $s{=}4$ & 16.6 & 9.66 & 9.7 & 17.16 & 1.20 \\
\bottomrule
\end{tabular}
}
\end{table}

\subsection{Speed of Collision Plans}
Table~\ref{tab:speedup} isolates the scenes that collide only under a
degradation, and compares their 5\,s progress with that of the clean plan of
the same scene. A positive change means the degraded plan covers more distance.

\begin{table}[h]
  \centering
  \caption{Scenes that collide only under the degradation. Faster: share whose
  plan covers more distance than the clean plan. $\Delta$Prog.\ in metres.}
  \label{tab:speedup}
  \resizebox{\columnwidth}{!}{
\begin{tabular}{lrrrr}
\toprule
Condition & New coll. & Faster (\%) & $\Delta$Prog.\ mean & median \\
\midrule
Glare $s{=}3$, on & 34 & 82.4 & +5.34 & +5.86 \\
Glare $s{=}5$, on & 44 & 86.4 & +5.07 & +5.41 \\
Occlusion $s{=}3$, on & 39 & 92.3 & +5.36 & +5.15 \\
Tunnel exit $s{=}5$, sudden & 27 & 88.9 & +2.38 & +2.56 \\
Fog $s{=}4$ & 42 & 90.5 & +4.77 & +3.68 \\
Rain $s{=}4$ & 30 & 76.7 & +4.82 & +3.18 \\
\bottomrule
\end{tabular}
}
\end{table}

\subsection{Breakdown by Scene Attributes}
Table~\ref{tab:splits} groups collision-rate changes by scene attribute.
The attributes are hazard class and distance, city, ego stopping status,
and navigation command. Each column is the change against the clean rate
of the same subset, in percentage points.

\begin{table*}[t]
  \centering
  \caption{Change in collision rate (points) against the clean run, by scene
  attribute. Clean: collision rate of the clean run in the subset (\%).}
  \label{tab:splits}
  \small
\begin{tabular}{llrrrrrrr}
\toprule
Split & Value & $n$ & Clean & Occl.\ on & Occl.\ off & Glare$_5$ on & Glare$_3$ off & Fog \\
\midrule
\multirow{3}{*}{class} & car & 426 & 22.5 & +5.4 & +0.5 & +4.9 & -0.2 & +4.0 \\
 & pedestrian & 170 & 5.3 & +5.9 & +1.8 & +6.5 & -1.8 & +5.3 \\
 & other & 92 & 6.5 & +0.0 & +0.0 & +1.1 & +1.1 & +2.2 \\
\midrule
\multirow{3}{*}{distance} & $<$10\,m & 69 & 11.6 & +7.2 & +0.0 & +7.2 & +1.4 & +5.8 \\
 & 10--20\,m & 311 & 28.6 & +8.4 & +2.3 & +7.7 & -1.0 & +8.0 \\
 & 20--35\,m & 308 & 4.5 & +0.6 & -0.6 & +1.3 & -0.3 & -0.3 \\
\midrule
\multirow{2}{*}{city} & Singapore & 425 & 12.9 & +3.5 & +0.9 & +4.9 & +0.7 & +4.2 \\
 & Boston & 263 & 21.3 & +6.8 & +0.4 & +4.6 & -2.3 & +3.8 \\
\midrule
\multirow{2}{*}{speed} & stopped & 247 & 37.2 & +8.1 & +1.6 & +6.5 & +0.0 & +6.5 \\
 & moving & 441 & 4.3 & +2.9 & +0.2 & +3.9 & -0.7 & +2.7 \\
\midrule
\multirow{3}{*}{command} & keep forward & 634 & 17.5 & +5.2 & +0.8 & +5.2 & -0.6 & +4.4 \\
 & turn left & 15 & 0.0 & +0.0 & +0.0 & +0.0 & +0.0 & +0.0 \\
 & turn right & 39 & 0.0 & +0.0 & +0.0 & +0.0 & +2.6 & +0.0 \\
\bottomrule
\end{tabular}

\end{table*}

\subsection{Validity of the Collision Check}
\label{app:collision}
The collision check of Section~\ref{sec:benchmark} ignores headings. We
repeat the check with oriented boxes, using the direction of travel of the plan for
the ego vehicle and the annotated heading of each hazard along its track. The
human trajectory collides in \gtOriented{}\% of the scenes under both checks.
Every conclusion of the diagnosis holds with oriented boxes, as shown in Table~\ref{tab:robust}.

The clean plans collide in \collClean{}\% of the scenes. These collisions have
one source. In \cleanFarther{}\% of these scenes the plan covers more distance than the
human driver. Most of the hazards are stationary cars ahead, and the first
contact occurs at a median of \cleanFirstMed{}\,s, after 3\,s in
\cleanAfterThree{}\% of cases. The planner extends its current speed and
reaches a car that the human driver stopped behind. This is consistent with the
dominance of the ego state in open-loop planning \citep{egostatus}. All results
are paired against the clean run of the same scene, so this baseline cancels in
every comparison.

Collisions within the first 3\,s are rare for every condition (Table~\ref{tab:robust}). The effect of a hidden hazard therefore appears in
the second half of the horizon. This timing is consistent with sustained planned speed. The plan maintains
speed and reaches the hazard only after several seconds.

\begin{table}[h]
  \centering
  \caption{Change in collision rate (points) against the clean run under four
  variants of the collision check, on the diagnosis scenes. $^*$: $p<0.05$,
  exact McNemar test.}
  \label{tab:robust}
  \resizebox{\columnwidth}{!}{
\begin{tabular}{lrrrrrrr}
\toprule
Check & Clean & Occl.\ on & Occl.\ off & Glare$_3$ on & Glare$_3$ off & Glare$_5$ on & Fog \\
\midrule
5\,s, axis-aligned & 16.1 & +4.8$^*$ & +0.7 & +3.8$^*$ & -0.4 & +4.8$^*$ & +4.1$^*$ \\
5\,s, oriented & 16.0 & +4.8$^*$ & +1.3 & +3.6$^*$ & -0.6 & +5.4$^*$ & +4.7$^*$ \\
3\,s, axis-aligned & 0.6 & +0.4 & +0.0 & +0.3 & +0.4 & +1.2$^*$ & +0.4 \\
3\,s, oriented & 0.6 & +0.4 & +0.0 & +0.4 & +0.3 & +1.5$^*$ & +0.9 \\
\bottomrule
\end{tabular}
}
\end{table}

\subsection{Noise Floor}
\model{} decodes action tokens by sampling at temperature 0.01 without a fixed
seed. Two clean runs of the same scene give the same collision outcome in all
but a few scenes (Table~\ref{tab:diag}, clean rerun). The final pose of the two
runs differs by about 0.1\,m on average, while the degradations shift the final pose by
several metres. The main degradation effects therefore exceed
the decoding noise.

\section{Comparison on the Development Scenes}
\label{app:compare}

\begin{table}[t]
  \centering
  \caption{Collision rate (\%) of \model{} with each training-free repair on
  \nRep{} scenes. Res./New: scenes whose collision the repair resolves /
  introduces, relative to no repair. $p$: exact McNemar test. Bold: lowest
  collision rate among deployable repairs. $^\dagger$Uses ground truth about
  which frames or pixels are degraded.}
  \label{tab:compare}
  \resizebox{\columnwidth}{!}{
\begin{tabular}{lrrrrrrrrr}
\toprule
 & \multicolumn{3}{c}{Occlusion, on} & \multicolumn{3}{c}{Glare $s{=}5$, on} & \multicolumn{3}{c}{Fog} \\
\cmidrule(lr){2-4} \cmidrule(lr){5-7} \cmidrule(lr){8-10}
Repair & Coll. & Res./New & $p$ & Coll. & Res./New & $p$ & Coll. & Res./New & $p$ \\
\midrule
No repair & 21.0 & -- & -- & 20.8 & -- & -- & 20.2 & -- & -- \\
\midrule
CLAHE & 19.5 & 19/10 & 0.136 & 21.7 & 11/16 & 0.442 & 19.5 & 20/16 & 0.618 \\
Dark channel prior & 19.8 & 24/16 & 0.268 & 19.0 & 21/10 & 0.071 & \textbf{15.8} & 36/10 & $<$0.001 \\
Inpainting (current frame) & 20.7 & 16/14 & 0.856 & 18.8 & 32/20 & 0.126 & 19.0 & 34/27 & 0.443 \\
Frame freeze (detector) & \textbf{15.8} & 44/13 & $<$0.001 & \textbf{15.8} & 44/14 & $<$0.001 & 17.3 & 31/14 & 0.016 \\
\method{}, region refill & 18.0 & 27/9 & 0.004 & 18.8 & 26/14 & 0.081 & 16.7 & 31/10 & 0.001 \\
\method{}, adaptive (ours) & 16.0 & 42/12 & $<$0.001 & 16.0 & 42/13 & $<$0.001 & 17.2 & 31/13 & 0.010 \\
\midrule
Frame freeze, known frames$^\dagger$ & 15.5 & 42/9 & $<$0.001 & 15.7 & 41/10 & $<$0.001 & 17.2 & 32/14 & 0.011 \\
\method{}, oracle mask$^\dagger$ & 17.3 & 29/7 & $<$0.001 & 16.3 & 38/11 & $<$0.001 & 17.7 & 33/18 & 0.049 \\
\bottomrule
\end{tabular}
}
\end{table}

Table~\ref{tab:compare} gives the full comparison discussed in Section~\ref{sec:compare}.

\section{Ablation of Region Refill}
\label{app:ablation}

\begin{table}[t]
  \centering
  \caption{Ablation of \method{} region refill on the same \nRep{} scenes.
  To isolate each component, all rows are applied to the frames known to be
  degraded. Res./New and $p$ are relative to no repair.}
  \label{tab:ablation}
  \resizebox{\columnwidth}{!}{
\begin{tabular}{lrrrrrrrrr}
\toprule
 & \multicolumn{3}{c}{Occlusion, on} & \multicolumn{3}{c}{Glare $s{=}5$, on} & \multicolumn{3}{c}{Fog} \\
\cmidrule(lr){2-4} \cmidrule(lr){5-7} \cmidrule(lr){8-10}
Repair & Coll. & Res./New & $p$ & Coll. & Res./New & $p$ & Coll. & Res./New & $p$ \\
\midrule
\method{} (full) & 18.2 & 26/9 & 0.006 & 19.0 & 24/13 & 0.099 & 16.8 & 30/10 & 0.002 \\
w/o alignment & \textbf{17.5} & 30/9 & 0.001 & \textbf{17.0} & 27/4 & $<$0.001 & 17.2 & 27/9 & 0.004 \\
saturation cue only & 20.3 & 10/6 & 0.454 & 17.2 & 42/20 & 0.007 & 20.7 & 7/10 & 0.629 \\
texture cue only & 18.5 & 28/13 & 0.028 & 19.1 & 28/17 & 0.135 & \textbf{16.5} & 32/10 & $<$0.001 \\
\midrule
oracle mask & 17.3 & 29/7 & $<$0.001 & 16.3 & 38/11 & $<$0.001 & 17.7 & 33/18 & 0.049 \\
frame freeze & 15.5 & 42/9 & $<$0.001 & 15.7 & 41/10 & $<$0.001 & 17.2 & 32/14 & 0.011 \\
\bottomrule
\end{tabular}
}
\end{table}

Table~\ref{tab:ablation} removes one component of region refill at a time.
Each cue covers different families. The saturation cue alone helps under glare
(\repGlFiveOnSat{}\%) but not under occlusion (\repOcOnSat{}\%), since a dark
occluder does not saturate. The texture cue alone helps under occlusion and fog
(\repOcOnTex{}\% and \repFogTex{}\%) but less under glare
(\repGlFiveOnTex{}\%). Removing the homography lowers the collision rate under
occlusion from \repOcOnTrf{}\% to \repOcOnNoAlign{}\% and under glare from
\repGlFiveOnTrf{}\% to \repGlFiveOnNoAlign{}\%. Over 2\,s the homography
corrects distant structure but stretches near objects and the image border.
The result is consistent with these artefacts costing the planner more than
the age of the reference.
The same holds for the whole frame: replacing the frame outright
(\repOcOnFreeze{}\%) does better than refilling a large region. This is the
reason for the adaptive switch.

\section{Cost on Clean Input}
\label{app:cleancost}

\begin{table}[t]
  \centering
  \caption{Effect of each repair when applied to clean input on \nRep{} scenes.}
  \label{tab:cleancost}
  \resizebox{\columnwidth}{!}{
\begin{tabular}{lrrrr}
\toprule
Applied to clean input & Coll. & New/Res. & $p$ & $\Delta$L2 \\
\midrule
None & 16.3 & -- & -- & -- \\
\method{}, adaptive & 14.8 & 5/14 & 0.064 & -0.02 \\
\method{}, region refill & 15.3 & 5/11 & 0.210 & -0.03 \\
Frame freeze (detector) & 17.0 & 19/15 & 0.608 & +0.07 \\
CLAHE & 15.7 & 13/17 & 0.585 & -0.02 \\
Dark channel prior & 15.5 & 19/24 & 0.542 & +0.03 \\
\bottomrule
\end{tabular}
}
\end{table}

A repair runs on every frame and must preserve performance on clean input.
Table~\ref{tab:cleancost} applies each repair to the clean frames. The
detector flags some area on most clean frames, mostly where the scene changed
between the reference and the current frame (Table~\ref{tab:detmisc}). Frame
freeze triggered by the detector therefore replaces many clean frames. Detector-triggered frame freeze is
the only repair that raises the clean collision rate, from \cleanNone{}\% to
\cleanDetFreeze{}\%, with \cleanNewDetFreeze{} new and \cleanResDetFreeze{}
resolved collisions. Adaptive \method{} gives \cleanAdapt{}\%
(\cleanNewAdapt{} new, \cleanResAdapt{} resolved). Under degradation the two
are close (Table~\ref{tab:compare}). Adaptive \method{} retains the collision reduction
without increasing the collision rate on clean input.

\section{Placement and Per-Planner Repair}
\label{app:onoffrepair}

\begin{table}[t]
  \centering
  \caption{Direct comparison of the on and off placements on the same scenes.
  $\Delta$Prog.\ and $\Delta$Gap: on minus off (m), with a Wilcoxon test.}
  \label{tab:onoff}
  \resizebox{\columnwidth}{!}{
\begin{tabular}{lrrrrrrr}
\toprule
Family & $n$ & Coll.\ on & Coll.\ off & New/Res. & $p$ & $\Delta$Prog.\ ($p$) & $\Delta$Gap \\
\midrule
Glare $s{=}3$ & 690 & 19.9 & 15.7 & 44/15 & $<$0.001 & -0.09 (0.499) & -0.10 \\
Occlusion $s{=}3$ & 691 & 20.8 & 16.8 & 37/9 & $<$0.001 & +0.66 ($<$0.001) & -0.19 \\
\bottomrule
\end{tabular}
}
\end{table}

\begin{table}[t]
  \centering
  \caption{Adaptive \method{} on each planner: collision rate (\%) without
  $\to$ with the repair, and change in progress (m). $^*$: $p<0.05$, McNemar.}
  \label{tab:modelsrepair}
  \resizebox{\columnwidth}{!}{
\begin{tabular}{lcccc}
\toprule
Planner & Occl.\ on & Glare$_5$ on & Fog & Clean \\
\midrule
AutoVLA & 21.0 $\to$ 16.0$^*$ (-1.0) & 20.9 $\to$ 16.0$^*$ (-1.4) & 20.2 $\to$ 17.2$^*$ (-0.2) & 16.4 $\to$ 14.9 (+0.1) \\
Qwen-Drive & 3.3 $\to$ 4.2 (+1.0) & 3.2 $\to$ 4.3 (+0.5) & 11.7 $\to$ 4.0$^*$ (-0.5) & 2.7 $\to$ 3.2 (-0.0) \\
Impromptu-VLA & 0.0 $\to$ 0.0 (-0.5) & 0.0 $\to$ 0.0 (-0.2) & 0.0 $\to$ 0.0 (-0.5) & 0.0 $\to$ 0.0 (+0.0) \\
\bottomrule
\end{tabular}
}
\end{table}

Table~\ref{tab:onoff} compares the on and off placements directly (Section~\ref{sec:diagnosis}). Table~\ref{tab:modelsrepair} applies adaptive \method{} to each planner (Section~\ref{sec:planners}).

\section*{Acknowledgment}
Figure icons are from Font Awesome Free, licensed under CC BY 4.0.

\bibliographystyle{template/IEEEtran}
\bibliography{references}

\end{document}